\pdfoutput=1

\documentclass[11pt]{article}

\usepackage[final]{acl}

\usepackage{times}
\usepackage{latexsym}

\usepackage[T1]{fontenc}

\usepackage[utf8]{inputenc}

\usepackage{microtype}

\usepackage{inconsolata}

\usepackage{hyperref}       
\usepackage{url}            
\usepackage{graphicx}
\usepackage{amsfonts}
\usepackage{amsmath}
\usepackage{amssymb}
\usepackage{mathtools}
\usepackage{bm}
\usepackage{bbm}
\usepackage{booktabs}
\usepackage{multirow}
\usepackage{multicol}
\usepackage{nicefrac}       
\usepackage{microtype}      
\usepackage{xcolor}         
\usepackage{adjustbox}
\usepackage{color}
\usepackage{colortbl}
\usepackage{cuted}

\title{Small Language Model as Data Prospector for Large Language Model}
\author{
\textbf{Shiwen Ni\textsuperscript{1}}\thanks{Equal contribution},
  \textbf{Haihong Wu\textsuperscript{1,2}}\footnotemark[1], \textbf{Di Yang\textsuperscript{1,2}},
      \textbf{Qiang Qu\textsuperscript{1}},
      \textbf{Hamid Alinejad-Rokny\textsuperscript{3}},
  \textbf{Min Yang\textsuperscript{1,4}}\thanks{Corresponding author}
\\
\\
  \textsuperscript{1}Shenzhen Key Laboratory for High Performance Data Mining,\\
  Shenzhen Institute of Advanced Technology, Chinese Academy of Sciences\\
  \textsuperscript{2}University of Science and Technology of China  ~~
  \textsuperscript{3}The University of New South Wales
  \\  
\textsuperscript{4}Shenzhen University of Advanced Technology
\\  
  {\{sw.ni, min.yang\}@siat.ac.cn}; \{haihongw, di-yang\}@mail.ustc.edu.cn}

\begin{document}
\maketitle
\begin{abstract}
The quality of instruction data directly affects the performance of fine-tuned Large Language Models (LLMs). Previously, \cite{li2023one} proposed \texttt{NUGGETS}, which identifies and selects high-quality quality data from a large dataset by identifying those individual instruction examples that can significantly improve the performance of different tasks after being learnt as one-shot instances. In this work, we propose \texttt{SuperNUGGETS}, an improved variant of \texttt{NUGGETS} optimised for efficiency and performance. Our \texttt{SuperNUGGETS} uses a small language model (SLM) instead of a large language model (LLM) to filter the data for outstanding one-shot instances and refines the predefined set of tests. The experimental results show that the performance of \texttt{SuperNUGGETS} only decreases by 1-2\% compared to \texttt{NUGGETS}, but the efficiency can be increased by a factor of 58. Compared to the original \texttt{NUGGETS}, our \texttt{SuperNUGGETS} has a higher utility value due to the significantly lower resource consumption. 
\end{abstract}

\section{Introduction}
Large Language Models (LLMs) have demonstrated excellent performance on a wide range of Natural Language Processing (NLP) tasks by scaling model size and datasets \cite{OpenAI2023GPT4TR, Anil2023PaLM2T,bai2023qwen,cheng2024towards} . Fine-tuning LLMs can further enhance the utility of these models by enabling them to better follow human instructions. This process usually involves supervised fine-tuning of input-output pairs, also known as instruction fine-tuning. This kind of fine-tuning not only awakens the knowledge acquired by the model during the pre-training phase, but also allows the model to interact with humans in a more natural conversational form.

\begin{figure}[t]
  \includegraphics[width=\columnwidth]{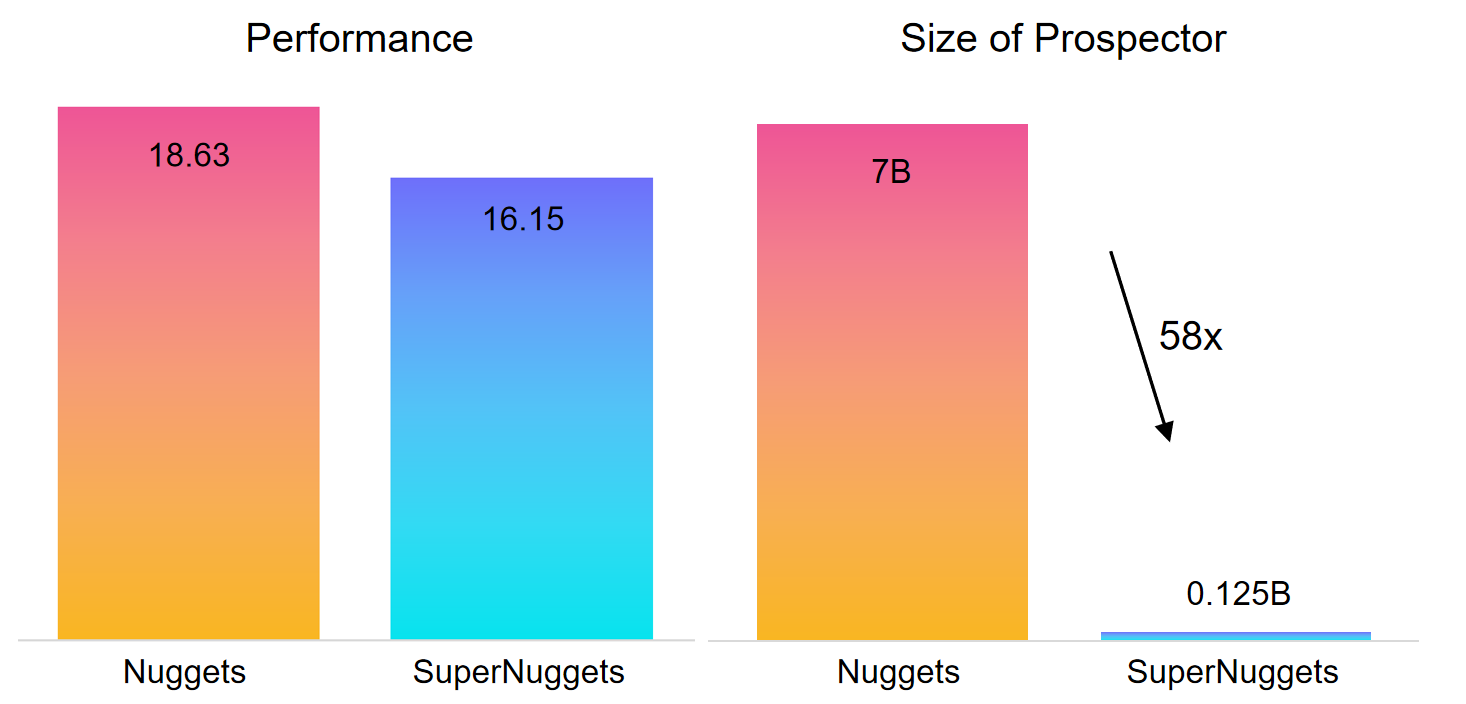}
  \caption{Comparison of \texttt{Nuggets} and Super\texttt{Nuggets} on the Alpaca-Eval benchmark.}
  \label{fig:experiments}
\end{figure}

Currently, much research \cite{chung2022scaling,wang2022super,wang2023far} is devoted to optimizing instruction fine-tuning by collecting larger, diverse, and complex datasets, often derived from open source data or expanded based on large language models. However, some recent studies \citep{bai2024coig,zhou2023lima,cao2023instruction} have shown that smaller but carefully selected high-quality datasets in the instruction fine-tuning phase can be more helpful in improving model performance. Performing simply the quantity of data while neglecting the quality of the data may lead to degradation of the performance of the model. It \citep{dai2022can} has been shown that context learning can be approximated as implicitly forward-propagating fine-tuning, whereas instruction fine-tuning is realized by back-propagation . Therefore, \cite{li2023one} proposes the \texttt{Nuggets} method, which predicts the effect of instruction fine-tuning by the performance of context learning.
\texttt{NUGGETS} utilizes one-shot learning to sift through large amounts of data to find high-quality instruction data. Specifically, if an instruction example can significantly improve the model's performance on a specific task, then it is an example worth training. If an example can have a positive impact on multiple examples, then it is an important instruction data. This is done by first identifying a predefined set of tasks containing multiple examples, and then using the remaining examples as a candidate set. An example from the candidate set is sequentially selected as a one-shot example for contextual learning and scored by observing its impact on the perplexity of the predefined examples. This score reflects the correlation between the predefined examples and the candidate examples and serves as a criterion for data selection. Since the \texttt{NUGGETS} method needs to calculate the one-shot score and zero-shot score for each piece of data, the computation is very large. Moreover, the set of predefined tasks in \texttt{NUGGETS} is obtained by random sampling, and has not been quality checked and filtered, which may contain noise and will inevitably contain some low-quality data, which will directly affect the correctness of the subsequent score calculation. To address the above shortcomings in the original \texttt{NUGGETS} approach, we propose \texttt{SuperNUGGETS}, which utilises SLM to identify high-quality one-shot instances and perform refinement based on quality and diversity on the predefined set of test tasks.

\begin{figure}[t]
  \centering
\includegraphics[width=1\linewidth]{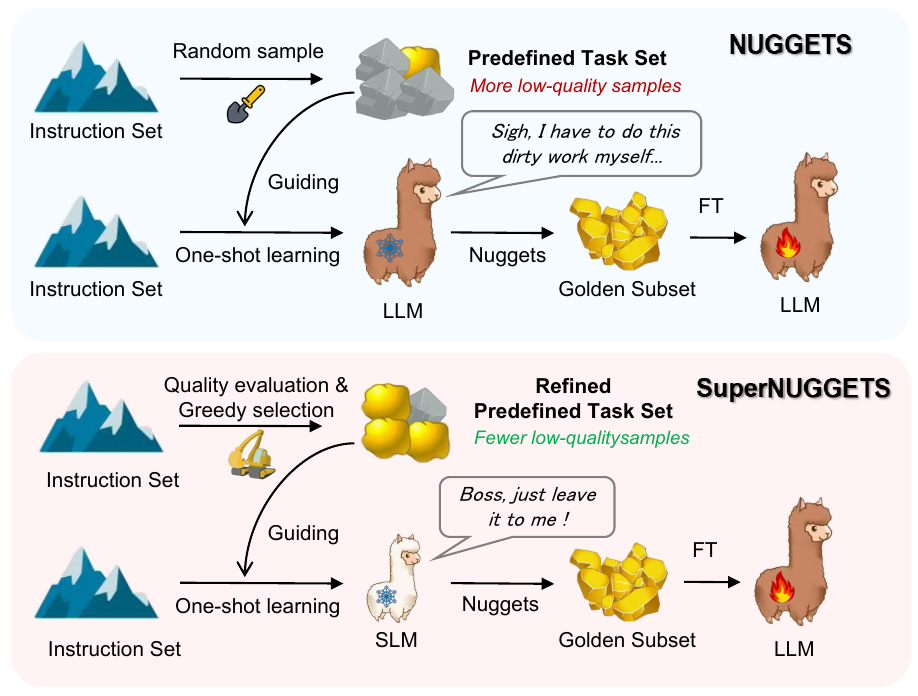}
  \caption{Comparison of \texttt{SuperNUGGETS} and \texttt{NUGGETS}.}
  \label{fig:experiments}
  \vspace{-1em}
\end{figure}

\section{SuperNUGGETS}
\textbf{Motivation} \texttt{NUGGETS} \citep{li2023one} utilizes one-shot learning to filter out high-quality instruction data from a large amount of instruction data, achieving excellent data prospecting results. However the original \texttt{NUGGETS} method requires calculating the one-shot score and zero-shot score for each piece of data, and the original size of the predefined task set is 1,000 pieces. To filter Alpaca's 52k pieces of data requires the model to inference a total of 52,002 (zero-shot) + [52,002 $\times$ 1,000] (one-shot) = 52,054,002 times. Using LLM to inference 104 million times is very time costing as well as a big resource drain. In addition, the predefined task set in the original \texttt{NUGGETS} method is obtained by random sampling, which will inevitably contain some low-quality data, which will directly affect the correctness of the subsequent calculation of the gold score. Therefore, to address the above limitations and problems, we propose \texttt{SuperNUGGETS}, an enhanced version of \texttt{NUGGETS}.
\vspace{-0.5em}
\subsection{Predefined Task Set Refinement}
The number of the alpaca dataset is 52,002, and our first step is to start with the reward model reward-model-deberta-v3-large-v2, to score the overall data. Then the top 10,000 data are filtered based on the score, of which the top 20 are taken separately as a high quality subset, this step is to ensure the high quality of the data. The second step encodes the first 20-10,000 data to obtain semantic vectors, which are clustered using the kcenter\_greedy algorithm. Specifically, an initial centroid is selected from the 20-10,000 dataset, usually the data point furthest from the other centroids. The data point furthest from the current set of centroids is then iteratively selected as the new centroid to ensure broader coverage. Finally the point furthest from the current set of centroids is iteratively selected, which ensures that the selected data is as dispersed as possible, covering all aspects of the instruction types, ensuring diversity and coverage of the selected instruction data. This step selects 80 examples from 20-1,000 data, and finally combines the 20 examples from the first step with the 80 examples from the second step to form a refined predefined test set containing 100 examples.

\begin{table*}[t]
\centering
\resizebox{0.93\textwidth}{!}{
\begin{tabular}{llccccccc}
\toprule	
\textbf{Data Prospector}            & \textbf{Data ratios}   & \multicolumn{1}{l}{\textbf{Helpful\_base}} & \multicolumn{1}{l}{\textbf{Koala}} & \multicolumn{1}{l}{\textbf{Self-instruct}} & \multicolumn{1}{l}{\textbf{Oasst}} & \multicolumn{1}{l}{\textbf{Vicuna}} & \multicolumn{1}{l}{\textbf{Length}} & \multicolumn{1}{l}{\textbf{Overall}} \\ \hline
/                          & 0\%           & 6.98                              & 10.26                     & 11.17                             & 9.92                      & 0.09                       & 1,593                      & 9.69                        \\
/                          & 100\% (full)  & 24.81                             & 18.59                     & 13.10                             & 24.47                     & 15.00                      & 357                        & \underline{18.51}                       \\ \hline
\multirow{6}{*}{Llama2-7B} & 1\% (top)     & 26.36                             & 14.74                     & 10.32                             & 22.34                     & 16.25                      & 434                        & 17.14                       \\
                           & 5\% (top)     & 37.98                             & 23.72                     & 18.65                             & 27.66                     & 16.25                      & 433                        & \textbf{24.47}                       \\
                           & 10\% (top)    & 24.03                             & 29.49                     & 17.86                             & 26.60                     & 20.00                      & 426                        & \textbf{23.29}                       \\
                           & 30\% (top)    & 26.36                             & 21.15                     & 16.67                             & 26.60                     & 17.50                      & 384                        & \textbf{21.37}                       \\
                           & 50\% (top)    & 18.60                             & 16.67                     & 14.68                             & 27.66                     & 13.75                      & 358                        & \textbf{18.63}                       \\
                           & 50\% (bottom) & 20.93                             & 15.38                     & 11.90                             & 19.15                     & 13.75                      & 331                        & 15.53                       \\ \hline
\multirow{6}{*}{Opt-350m}  & 1\% (top)     & 21.71                             & 13.46                     & 10.32                             & 22.34                     & 11.25                      & 439                        & 15.65                       \\
                           & 5\% (top)     & 38.76                             & 23.08                     & 12.30                             & 31.38                     & 21.25                      & 490                        & \textbf{23.98}                       \\
                           & 10\% (top)    & 31.01                             & 21.79                     & 14.29                             & 28.19                     & 18.75                      & 491                        & \textbf{21.99}                       \\
                           & 30\% (top)    & 23.26                             & 19.23                     & 15.08                             & 30.85                     & 17.50                      & 405                        & \textbf{20.99}                       \\
                           & 50\% (top)    & 23.26                             & 23.72                     & 15.87                             & 29.26                     & 15.00                      & 393                        & \textbf{21.61}                       \\
                           & 50\% (bottom) & 13.95                             & 13.46                     & 12.70                             & 22.87                     & 10.00                      & 295                        & 14.91                       \\ \hline
\multirow{6}{*}{Opt-125m}  & 1\% (top)     & 20.16                             & 13.46                     & 11.90                             & 21.28                     & 16.25                      & 421                        & 16.15                       \\
                           & 5\% (top)     & 32.56                             & 20.51                     & 14.68                             & 27.13                     & 20.00                      & 406                        & \textbf{22.11}                       \\
                           & 10\% (top)    & 27.91                             & 18.59                     & 13.89                             & 28.19                     & 25.00                      & 405                        & \textbf{21.37}                       \\
                           & 30\% (top)    & 24.81                             & 18.59                     & 19.44                             & 29.26                     & 26.25                      & 393                        & \textbf{23.04}                       \\
                           & 50\% (top)    & 21.71                             & 20.51                     & 16.67                             & 25.53                     & 16.25                      & 385                        & \textbf{20.12}                       \\
                           & 50\% (bottom) & 10.08                             & 12.82                     & 9.92                              & 18.62&6.25 	&295 	&11.86     \\
\bottomrule      
\end{tabular}}
\caption{The win\_rate results of models fine-tuned using different data under the Alpaca-Eval benchmark.}
\label{t1}
\end{table*}
\vspace{-0.5em}
\subsection{SLM as Instruction Data Prospector}
With an instruction tuning dataset $\mathcal{D}$, we aim to identify a set of examples $\mathcal{D}_{\text{gold}}$ that are most closely aligned with the golden instructions.
Like the original \texttt{NUGGETS} \citep{li2023one} method, we first need to calculate the zero-shot score for refined predefined task set. The predefined test set after the previous refinement encompasses a variety of $m$ tasks, where each task is structured as $\{\text{Task~(T)}, \text{Answer~(A)}\}$. Each token in $\text{Task}$ or $\text{Answer}$ is denoted as ${tk}^{\text{T}}_i$ or ${tk}^{\text{A}}_i$. Let $\textsf{SLM}$ denote the instruction data prospector we use. For the $j$-th task represented by ${T}_j$, the probability of zero-shot inference by the data prospector can be calculated by continuously predicting the next tokens based on the given task and the preceding words:
\begin{equation}
\begin{split}
        s^j_{\text{zero}} =& \frac{1}{L}\sum_{i=1}^{L}\log p({tk}^{\text{A}_j}_i | InP_{zero};\textsf{SLM}), \\
        InP_{zero} &= [{T}_j, {tk}_1^{\text{A}_j},{tk}_2^{\text{A}_j},\ldots,{tk}_{i-1}^{\text{A}_j}],
\end{split}
\end{equation}
where $L$ is the number of tokens of the ground-truth answer $\text{A}$. The score ${s}^j_{\text{zero}}$ is used to denote the competence level of the $\textsf{SLM}$ on the $j$th task. A higher ${s}^j_{\text{zero}}$ denotes superior model performance on the $j$-th task, whereas a lower ${s}^j_{\text{zero}}$ implies inferior performance. Therefore, we can acquire the data prospector's performance across $m$ tasks as:
\begin{equation}
     \boldsymbol {S}_{\text{zero}} = [{s}^1_{\text{zero}}, {s}^2_{\text{zero}},\ldots, {s}^{m-1}_{\text{zero}}, {s}^{m}_{\text{zero}}].
\end{equation}

For each example $\mathbf{z}_k=\{\text{IQ}_k,\text{IA}_k\}$, we initially perform one-shot learning on the base model using that specific example. 
Here, $\text{IQ}_k$ denotes the question associated with the $k$-th example $\mathbf{z}_k\in \mathcal{D}$, while $\text{IA}_k$ signifies its corresponding answer.
Subsequently, we employ the model with in-context learning to conduct another round of testing on the tasks within the predefined task set. That is,
\begin{equation}
    \begin{split}
        {s}^j_{\text{one}}(\mathbf{z}_k)  
    &= \frac{1}{L}\sum_{i=1}^{L}\log p({w}^{\text{A}_j}_i |InP_{one}; \texttt{SLM}), \\
    InP_{one} &=[{T}_j, \underbrace{(\text{IQ}_k,\text{IA}_k)}_{\textit{One-Shot Prompt}}, {w}^{\text{A}_j}_1,{w}^{\text{A}_j}_2,\ldots,{w}^{\text{A}_j}_{i-1}], 
    \end{split}
    \label{eq3}
\end{equation}
where $(\text{IQ}_k, \text{IA}_k)$ can be considered \textit{one-shot prompt}. 
Similarly, we can obtain the performance of the model after implicit fine-tuning across $m$ different tasks:
\begin{equation}
     \boldsymbol {S}^k_{\text{one}} = [{s}^1_{\text{one}}(\mathbf{z}_k), {s}^2_{\text{one}}(\mathbf{z}_k),\ldots, {s}^m_{\text{one}}(\mathbf{z}_k)].
\end{equation}
We use \textbf{Golden Score~(GS)} to reflect the score of our data prospector $\textsf{SLM}$ for that instruction data. The GS of the example $\mathbf{z}_k$ is calculated as 
\begin{equation}
    \text{GS}(\mathbf{z}_k) = \frac{1}{m}\sum_{i=1}^m \mathbbm{I}\left[{s}^i_{\text{one}}(\mathbf{z}_k)> {s}^i_{\text{zero}}\right]\  \in \  [0,1],
\end{equation}
where $\mathbbm{I}[\cdot]$ is the indicator function. The GS measures the increment of performance improvement of the model after one-shot learning through the given instruction.
Finally, we use GS to filter the data to get the top n\% of datasets $\mathcal{D}^{n \%}_{\text{gold}}$ with the highest GS as appropriate. Using SLM prospecting to get $\mathcal{D}^{n \%}_{\text{gold}}$ can be used to fine-tune the LLM.

\begin{table*}[t]
\centering
\resizebox{0.96\textwidth}{!}{
\begin{tabular}{llcccccc}
\toprule
\textbf{Data Prospector}            & \textbf{Predefined test sets} & \textbf{top 1\%} & \textbf{top 5\%} & \textbf{top 10\%} & \textbf{top 30\%} & \textbf{top 50\%} & \textbf{bottom 50\%} \\ \hline
\multirow{3}{*}{Llama2-7B} & 100 (Refined)        & 17.14   & 24.47   & 23.29    & 21.37    & 18.63    & 15.53       \\
                           & 100 (Random)         & 12.11   & 16.27   & 18.51    & 17.76    & 16.27    & 19.19       \\
                           & 1000 (Random)        & 18.63   & 21.49   & 23.79    & 21.37    & 19.32    & 17.14       \\ \hline
\multirow{3}{*}{Opt-350m}  & 100 (Refined)        & 15.65   & 23.98   & 21.99    & 20.99    & 21.61    & 14.91       \\
                           & 100 (Random)         & 15.28   & 20.56   & 18.14    & 17.45   & 19.57    & 17.70       \\
                           & 1000 (Random)        & 16.15   & 24.47   & 23.17    & 25.47    & 20.68    & 15.28       \\ \hline
\multirow{3}{*}{Opt-125m}  & 100 (Refined)        & 16.15   & 22.11   & 21.37    & 23.04    & 20.12    & 11.86       \\
                           & 100 (Random)         & 12.11   & 19.38   & 20.62    & 21.37    & 18.32    & 16.71       \\
                           & 1000 (Random)        & 13.29   & 20.56   & 20.62    & 22.86    & 20.56    & 15.28     \\
                           \bottomrule      
\end{tabular}}
\caption{Ablation study of predefined task set refinement.}
\label{t2}
\end{table*}

\begin{table}[t]
\centering
\resizebox{0.47\textwidth}{!}{
\begin{tabular}{l|ccc}
\toprule
Data Prospector     & \multicolumn{1}{l}{Llama2-7B} & \multicolumn{1}{l}{Opt-350m} & \multicolumn{1}{l}{Opt-125m} \\ \hline
Llama2-7B & 1                             & 0.701                        & 0.653                        \\
Opt-350m  & 0.701                         & 1                            & 0.786                         \\
Opt-125m  & 0.653                         & 0.786                         & 1   \\
\bottomrule    
\end{tabular}}
\caption{Percentage of identical data between the top 30\% of data screened by different data prospectors.}
\label{t3}
\end{table}

\section{Experiment}
\subsection{Experimental Setup}
As with \citep{li2023one}, we chose Alpaca as the instruction dataset to be used for data filtering. This dataset is pivotal within the open-source sphere for the purpose of instruction tuning. It was created using the self-instruct~\cite{wang2022self} technique, which extracts instruction data from text-davinci-003. The dataset's effectiveness in refining the LLaMA model has catalyzed a wave of research into the realm of instruction fine-tuning~\cite{li2023m,ji2023exploring,xu2023baize}.

Same as the original \texttt{NUGGETS} \citep{li2023one}, we compare the responses generated by the model with those generated by the davincici -003 model, using the well-established \textbf{Alpaca-Eval} dataset \cite{alpaca_eval}. This dataset uses ‘win\_rate’ as the evaluation metric. 
In our experiments, we use three models, Opt-125m, Opt-350m, and Llama2-7B, respectively, as data Prospector. We specify the Llama2-7B model as the base model for generation fine-tuning. In the model fine-tuning phase, we use an Adam optimiser with a learning rate of $2\times 10^{-5}$, a learning rate of 2e-5, a batch size of 16, a warmup\_ratio of 0.03, and an epoch of 3. In the subsequent model evaluation phase, we use the gpt-4o-mini for the measurement.
\subsection{Experimental Results}
As shown in Table \ref{t1}, we use Opt-125m, Opt-350m, and Llama2-7B as data prospectors, respectively, and the predefined test set is the refined 100 data. The results of model performance over using 100\% data (52,002) fine-tuning are bolded in the table. From the experimental results, it is evident that our \texttt{SuperNUGGETS} filtered data using only the top 5\% exceeds the effect of fine-tuning the model out of 100\% of the data. We found that the model trained on top 5\% of the data obtained using Opt-350m (20 times smaller than Llama2-7B) as the data prospector also achieves a score of 23.98, which is much higher than the model fine-tuned on 100\% of the full amount of data. Even the model trained with TOP 5\% of the data obtained using Opt-125m (56 times smaller than Llama2-7B) as the data Prospector achieves a score of 22.11, which is much higher than the model fine-tuned with 100\% of the full amount of data. All three models Opt-125m, Opt-350m, and Llama2-7B screened the top 50\% of the data better than the bottom 50\% of the data, which demonstrates the effectiveness of our \texttt{SuperNUGGETS}. As shown in Table \ref{t3}, we find that the top 30\% of the data screened by the three sizes of prospectors are very similar, which also indicates that SLM is an alternative to LLM as a data prospector. Case studies are in the appendix \ref{appendix_a}.
\section{Ablation Study}
While the original \texttt{NUGGETS} used 1,000 random data as a predefined task test set, our \texttt{SuperNUGGETS} uses a refined set of 100 data, which makes the number of computations 10 times smaller. As shown in Table \ref{t2}, using the refined 100 data as the predefined task test set is far better than randomly selecting 100 data, regardless of which model the data prospector is. We found that the effect of the refined 100 data was even similar to that of the randomly filtered 1,000 data. The above experimental results illustrate the validity letter of our refinement of the predefined task test set.

\section{Conclusion}
Previously, \cite{li2023one} proposed \texttt{NUGGETS}, which identifies and selects high-quality data from large datasets through the effect of one-shot learning. In this work, we propose \texttt{SuperNUGGETS}, which is an \texttt{NUGGETS} improved variant optimized for efficiency and performance. Our \texttt{SuperNUGGETS} uses a small language model (SLM) instead of a large language model (LLM) to filter unprocessed single instance data and refines a predefined test set. Experimental results show that \texttt{SuperNUGGETS} is only 1-2\% less performant than \texttt{NUGGETS}, but 58 times more efficient. Compared to the original \texttt{NUGGETS}, our \texttt{SuperNUGGETS} has a much higher utility value because of the significantly lower resource consumption. 

\section*{Limitations}
Due to funding and resource constraints, full-parameter fine-tuning was not carried out for models at scales above 7B. The performance of the filtered high-quality data on larger scale models is unknown.


\bibliography{custom}

\begin{thebibliography}{17}
\providecommand{\natexlab}[1]{#1}

\bibitem[{Bai et~al.(2023)Bai, Bai, Chu, Cui, Dang, Deng, Fan, Ge, Han, Huang
  et~al.}]{bai2023qwen}
Jinze Bai, Shuai Bai, Yunfei Chu, Zeyu Cui, Kai Dang, Xiaodong Deng, Yang Fan,
  Wenbin Ge, Yu~Han, Fei Huang, et~al. 2023.
\newblock Qwen technical report.
\newblock \emph{arXiv preprint arXiv:2309.16609}.

\bibitem[{Bai et~al.(2024)Bai, Du, Liang, Jin, Liu, Zhou, Zheng, Zhang, Ma,
  Wang et~al.}]{bai2024coig}
Yuelin Bai, Xinrun Du, Yiming Liang, Yonggang Jin, Ziqiang Liu, Junting Zhou,
  Tianyu Zheng, Xincheng Zhang, Nuo Ma, Zekun Wang, et~al. 2024.
\newblock Coig-cqia: Quality is all you need for chinese instruction
  fine-tuning.
\newblock \emph{arXiv preprint arXiv:2403.18058}.

\bibitem[{Cao et~al.(2023)Cao, Kang, and Sun}]{cao2023instruction}
Yihan Cao, Yanbin Kang, and Lichao Sun. 2023.
\newblock Instruction mining: High-quality instruction data selection for large
  language models.
\newblock \emph{arXiv preprint arXiv:2307.06290}.

\bibitem[{Cheng et~al.(2024)Cheng, Zhu, Li, Li, Zhuang, and
  Zou}]{cheng2024towards}
Xuxin Cheng, Zhihong Zhu, Hongxiang Li, Yaowei Li, Xianwei Zhuang, and Yuexian
  Zou. 2024.
\newblock Towards multi-intent spoken language understanding via hierarchical
  attention and optimal transport.
\newblock In \emph{Proceedings of the AAAI Conference on Artificial
  Intelligence}, volume~38, pages 17844--17852.

\bibitem[{Chung et~al.(2022)Chung, Hou, Longpre, Zoph, Tay, Fedus, Li, Wang,
  Dehghani, Brahma et~al.}]{chung2022scaling}
Hyung~Won Chung, Le~Hou, Shayne Longpre, Barret Zoph, Yi~Tay, William Fedus,
  Eric Li, Xuezhi Wang, Mostafa Dehghani, Siddhartha Brahma, et~al. 2022.
\newblock Scaling instruction-finetuned language models.
\newblock \emph{arXiv preprint arXiv:2210.11416}.

\bibitem[{Dai et~al.(2022)Dai, Sun, Dong, Hao, Sui, and Wei}]{dai2022can}
Damai Dai, Yutao Sun, Li~Dong, Yaru Hao, Zhifang Sui, and Furu Wei. 2022.
\newblock Why can gpt learn in-context? language models secretly perform
  gradient descent as meta optimizers.
\newblock \emph{arXiv preprint arXiv:2212.10559}.

\bibitem[{Google(2023)}]{Anil2023PaLM2T}
Google. 2023.
\newblock Palm 2 technical report.
\newblock \emph{arXiv preprint arXiv:2305.10403}.

\bibitem[{Ji et~al.(2023)Ji, Deng, Gong, Peng, Niu, Zhang, Ma, and
  Li}]{ji2023exploring}
Yunjie Ji, Yong Deng, Yan Gong, Yiping Peng, Qiang Niu, Lei Zhang, Baochang Ma,
  and Xiangang Li. 2023.
\newblock Exploring the impact of instruction data scaling on large language
  models: An empirical study on real-world use cases.
\newblock \emph{arXiv preprint arXiv:2303.14742}.

\bibitem[{Li et~al.(2023{\natexlab{a}})Li, Yin, Li, Chen, Wang, Ren, Li, Yang,
  Xu, Sun et~al.}]{li2023m}
Lei Li, Yuwei Yin, Shicheng Li, Liang Chen, Peiyi Wang, Shuhuai Ren, Mukai Li,
  Yazheng Yang, Jingjing Xu, Xu~Sun, et~al. 2023{\natexlab{a}}.
\newblock M3it: A large-scale dataset towards multi-modal multilingual
  instruction tuning.
\newblock \emph{arXiv preprint arXiv:2306.04387}.

\bibitem[{Li et~al.(2023{\natexlab{b}})Li, Zhang, Dubois, Taori, Gulrajani,
  Guestrin, Liang, and Hashimoto}]{alpaca_eval}
Xuechen Li, Tianyi Zhang, Yann Dubois, Rohan Taori, Ishaan Gulrajani, Carlos
  Guestrin, Percy Liang, and Tatsunori~B. Hashimoto. 2023{\natexlab{b}}.
\newblock Alpacaeval: An automatic evaluator of instruction-following models.
\newblock \url{https://github.com/tatsu-lab/alpaca_eval}.

\bibitem[{Li et~al.(2023{\natexlab{c}})Li, Hui, Xia, Yang, Yang, Zhang, Si,
  Liu, Liu, Huang et~al.}]{li2023one}
Yunshui Li, Binyuan Hui, Xiaobo Xia, Jiaxi Yang, Min Yang, Lei Zhang, Shuzheng
  Si, Junhao Liu, Tongliang Liu, Fei Huang, et~al. 2023{\natexlab{c}}.
\newblock One shot learning as instruction data prospector for large language
  models.
\newblock \emph{arXiv preprint arXiv:2312.10302}.

\bibitem[{OpenAI(2023)}]{OpenAI2023GPT4TR}
OpenAI. 2023.
\newblock Gpt-4 technical report.
\newblock \emph{arXiv preprint arXiv:2303.08774}.

\bibitem[{Wang et~al.(2023)Wang, Ivison, Dasigi, Hessel, Khot, Chandu, Wadden,
  MacMillan, Smith, Beltagy et~al.}]{wang2023far}
Yizhong Wang, Hamish Ivison, Pradeep Dasigi, Jack Hessel, Tushar Khot,
  Khyathi~Raghavi Chandu, David Wadden, Kelsey MacMillan, Noah~A Smith,
  Iz~Beltagy, et~al. 2023.
\newblock How far can camels go? exploring the state of instruction tuning on
  open resources.
\newblock \emph{arXiv preprint arXiv:2306.04751}.

\bibitem[{Wang et~al.(2022{\natexlab{a}})Wang, Kordi, Mishra, Liu, Smith,
  Khashabi, and Hajishirzi}]{wang2022self}
Yizhong Wang, Yeganeh Kordi, Swaroop Mishra, Alisa Liu, Noah~A Smith, Daniel
  Khashabi, and Hannaneh Hajishirzi. 2022{\natexlab{a}}.
\newblock Self-instruct: Aligning language model with self generated
  instructions.
\newblock \emph{arXiv preprint arXiv:2212.10560}.

\bibitem[{Wang et~al.(2022{\natexlab{b}})Wang, Mishra, Alipoormolabashi, Kordi,
  Mirzaei, Naik, Ashok, Dhanasekaran, Arunkumar, Stap et~al.}]{wang2022super}
Yizhong Wang, Swaroop Mishra, Pegah Alipoormolabashi, Yeganeh Kordi, Amirreza
  Mirzaei, Atharva Naik, Arjun Ashok, Arut~Selvan Dhanasekaran, Anjana
  Arunkumar, David Stap, et~al. 2022{\natexlab{b}}.
\newblock Super-naturalinstructions: Generalization via declarative
  instructions on 1600+ nlp tasks.
\newblock In \emph{EMNLP}, pages 5085--5109.

\bibitem[{Xu et~al.(2023)Xu, Guo, Duan, and McAuley}]{xu2023baize}
Canwen Xu, Daya Guo, Nan Duan, and Julian McAuley. 2023.
\newblock Baize: An open-source chat model with parameter-efficient tuning on
  self-chat data.
\newblock \emph{arXiv preprint arXiv:2304.01196}.

\bibitem[{Zhou et~al.(2023)Zhou, Liu, Xu, Iyer, Sun, Mao, Ma, Efrat, Yu, Yu
  et~al.}]{zhou2023lima}
Chunting Zhou, Pengfei Liu, Puxin Xu, Srini Iyer, Jiao Sun, Yuning Mao, Xuezhe
  Ma, Avia Efrat, Ping Yu, Lili Yu, et~al. 2023.
\newblock Lima: Less is more for alignment.
\newblock \emph{arXiv preprint arXiv:2305.11206}.

\end{thebibliography}

\appendix
\section{Case Study}
\label{appendix_a}
\begin{figure*}[t]
    \centering
    \includegraphics[width=\textwidth]{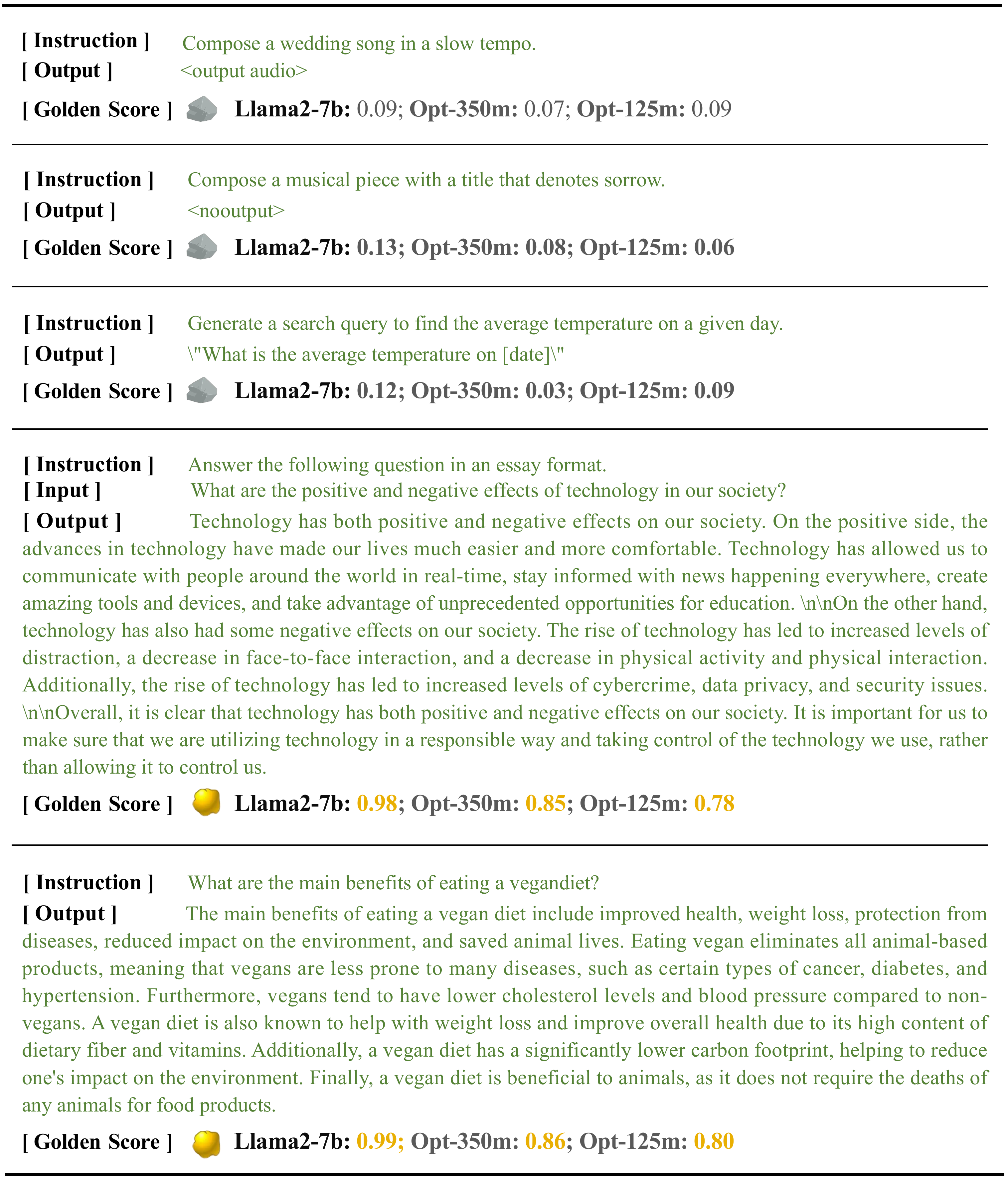}
    \caption{Examples of instructions and their corresponding golden scores.}
    \label{fig_case}
\end{figure*}

To qualitatively evaluate \texttt{SuperNUGGETS}, we also selected some example instructions from the Alpaca dataset for a case study, as shown in Figure \ref{fig_case}. We observe that instructions with very short and meaningless outputs give low gold scores for all three different sizes of data prospectors. In contrast, instructions with high gold scores are usually linguistically fluent, logically well thought out, have complete output, and are oriented towards helping humans solve problems.

\end{document}